\documentclass[letterpaper, 10 pt, conference]{ieeeconf}
\IEEEoverridecommandlockouts    
\usepackage{graphics} 
\usepackage{epsfig} 
\usepackage{amssymb}  

\usepackage{multirow}
\usepackage{subfigure}
\usepackage{bm}
\usepackage{float}
\usepackage{array}
\usepackage{amsmath}
\usepackage{booktabs,multirow}
\usepackage{makecell}
\usepackage{color}
\usepackage[cmyk]{xcolor}

\makeatletter
\usepackage{xspace}
\DeclareRobustCommand\onedot{\futurelet\@let@token\@onedot}
\def\@onedot{\ifx\@let@token.\else.\null\fi\xspace}

\def\etal{{et al}\onedot}
\makeatother

\def\etalcite#1{\etal~\cite{#1}}

\title{\LARGE \bf 
A Novel Multi-Gait Strategy for Stable and Efficient \\ Quadruped Robot Locomotion
}

\author{Daoxun Zhang \and Xieyuanli Chen \and Zhengyu Zhong \and Ming Xu \and Zhiqiang Zheng \and Huimin Lu
\thanks{All authors are with the College of Intelligence Science and Technology, National University of Defense Technology, China. {\tt\small e-mail: lhmnew@nudt.edu.cn}}
\thanks{This work was supported in part by the National Science Foundation of China under Grant U1913202, as well as Major Project of Natural Science Foundation of Hunan Province under Grant 2021JC0004.}%
}

\begin{document}

\maketitle
\thispagestyle{empty}
\pagestyle{empty}

\begin{abstract}
Taking inspiration from the natural gait transition mechanism of quadrupeds, devising a good gait transition strategy is important for quadruped robots to achieve energy-efficient locomotion on various terrains and velocities. While previous studies have recognized that gait patterns linked to velocities impact two key factors, the Cost of Transport (CoT) and the stability of robot locomotion, only a limited number of studies have effectively combined these factors to design a mechanism that ensures both efficiency and stability in quadruped robot locomotion. 
In this paper, we propose a multi-gait selection and transition strategy to achieve stable and efficient locomotion across different terrains. Our strategy starts by establishing a gait mapping considering both CoT and locomotion stability to guide the gait selection process during locomotion. Then, we achieve gait switching in time by introducing affine transformations for gait parameters and a designed finite state machine to build the switching order.
Comprehensive experiments have been conducted on using our strategy with changing terrains and velocities, and the results indicate that our proposed strategy outperforms baseline methods in achieving simultaneous efficiency in locomotion by considering CoT and stability.
\end{abstract}

\section{Introduction}
In the last decade, legged robots have taken a rapid development and been applied in diverse occasions due to their exceptional mobility and adaptability~\cite{ref-sheng2022bio, bledt2018cheetah,ref-nature, ref-fast, ref-ETH, ref-iit}. Quadruped robots combine the electromechanical system with a bio-inspired mechanism to overcome the flaws of regular wheeled robots, including traversing steps and maintaining balance on rugged terrains. However, people often ignore the benefits of quadruped robots' bio-inspired parts, for example, the effect of various gaits on economic locomotion.
In nature, quadruped animals, such as canids or felines, will modify their gait patterns and adjust the stride length when their body velocity changes. Specifically, they use the running or galloping gait for high-velocity motion, like chasing or preying, and deploy pace or trot gait for low-velocity motion, like strolling or searching. 
Similarly, many works have revealed that legged animals and robots will deploy different gait at various velocities in different environments to minimize energy consumption and maintain the stability of locomotion~\cite{intro1,intro2,intro3,ref-1986energy,ref-hoyt1981gait}. Therefore, enabling quadruped robots to adeptly choose and switch between gaits in response to changes in velocity and terrain holds significant potential for achieving energy-efficient locomotion.
\begin{figure}
	\centering
	\includegraphics[width=\linewidth]{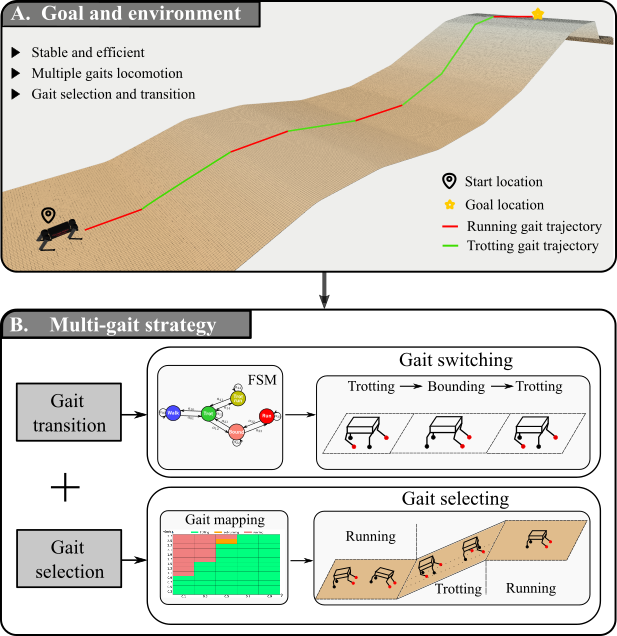}
	\caption{Illustration of the proposed strategy to achieve gait selecting and switching of quadruped robots. (A) shows the goal and working environment of a quadruped robot with different terrains. (B) shows the multi-gait strategy, including a gait transition between five gait patterns and a gait selection on two different terrains; the same color of feet represents the related legs swing simultaneously.}
	\label{total_figure}
 \vspace{-0.5cm}
\end{figure}

 Most research about quadruped robots focused on stability and versatility, relying on a predefined gait pattern~\cite{ref-robust, ref-ctdi2018dynamic,ref-design}. These studies ignored the benefit of gait selection and switching and limited the robot's energy utilization to a low level. 
Therefore, some research groups started to use the optimization on the gait selection of quadruped robots~\cite{5Automatic_gait_selection, 6Selecting_gaits} and built a gait mapping from task conditions to the optimal gaits, improving the quadruped robot's autonomous locomotion in different environments. 

Besides, in order to implement the suitable gait in the real world, we still should realize the gait switching in real-time when the robot is in an autonomous locomotion mode. The methods to achieve the gait switching can be categorized into two ways: one is to define the related gait parameters for each leg, which can be altered to realize different gaits~\cite{APS, 2bio_gait_transition, 3Terrain_adaptive}; the other is to embed the central pattern generator (CPG) in legs, coupling into a network which can achieve different gait patterns and realize the smooth gait switching~\cite{1CPG_gait_transition, 4Autonomous_gait_transition,santos2011gait,owaki2017quadruped}. Most gait transition methods based on CPG use the Hopf oscillator to output the joint control variables directly, which is efficient and useful. However, the Hopf-based CPG method can only output two variables that restrict the leg's joints into two, then limit the quadruped robot's active joints into 8, sacrificing the robot's versatility. Except for the methods mentioned above, machine learning and other intelligent algorithms can also be used to build a gait transition mechanism and achieve smooth switching, while these methods only consider the energy effect on different gaits emergence~\cite{ref-shao2021learning, ref-learning1,ref-learning2}.

In this paper, we introduce a multi-gait strategy for quadruped robots, as shown in Fig.~\ref{total_figure}. Our strategy first builds a mechanism for gait switching through the linear alteration of defined gait parameters and set switching sequence using a finite state machine. Then, the efficiency index (CoT) and the stable index (STB) are introduced to represent energy consumption and locomotion stability, respectively. Hundreds of tests' STB and CoT values are collected and merged to fit a gait mapping, and the most suitable gait for different terrains at corresponding speeds is obtained by querying the mapping table. Finally, this multi-gait strategy for selecting and transitioning between gaits is applied in real-time for quadruped robots running at different terrains.

In summary, we make three key claims: our approach is able to (i)~ realize the smooth switching among five distinct gaits and achieve highly dynamic gait motion; (ii)~construct a mapping mechanism from energy consumption, velocities, and stability to the gait patterns on different terrains through the evaluated information fusion; (iii)~build an evaluation mechanism based on energy efficiency and locomotion stability, and demonstrate the practical applicability of the gait selection theory from biology to quadrupedal robots.


\section{Related Work}
The gait switching of quadruped robots has been one of the important issues in the legged robots area since the natural gait transition mechanism has been found. Most researchers are dedicated to realizing quadruped robots' smooth and rapid gait transition to improve locomotion stability and robustness. Shang~\etalcite{1CPG_gait_transition} propose a Hopf oscillator-based CPG controller to generate different gait patterns directly and achieve smooth gait transition. Koo~\etalcite{2bio_gait_transition} design a quasi-static gait transition control strategy by defining the CASE and PHASE information to formulate a new footstep sequence, which considers the variation of CoM (Center of Moment) and focuses on the stable control. Furthermore, to realize the quadruped's stable locomotion on different terrains, a gait transition pattern using model predictive control is proposed~\cite{3Terrain_adaptive}, which can accomplish the switch between bounding and trotting by prolonging the period of stand phase and enhance the stability of the robot. Before the gait transition, finding an optimal gait for quadruped robots adapted to different speeds and tasks is necessary. For instance, Wang~\etalcite{5Automatic_gait_selection} proposes an automatic gait selection based on the idea of building a map from tasks to gait patterns. It utilizes mixed-integer nonlinear programming (MINLP) to generate the map and realize gait selection with supervised learning. However, all the works mentioned above focus on the stability control of the robot and are rarely concerned about energy efficiency.

In the field of robotics research, energy efficiency always plays an important role and should be considered primarily. A robot with high energy efficiency will take great advantage in many situations, such as exploring more areas, traveling longer distances, improving detection time, etc. Therefore, realizing efficient locomotion on quadruped robots will be necessary and valuable. Before that, the relevancy between locomotion and energy consumption should be known well. Prampero~\cite{ref-1986energy} shows that humans transition from walking to running to minimize their CoT. Hoyt and Taylor~\cite{ref-hoyt1981gait} show that, for the same reason, horses change from walking to trotting and from trotting to galloping as their velocity increases. 
Therefore, it can be realized that animals will achieve the most efficient gait instinctively with their locomotion velocity varying. Based on this biological principle, many scientists develop research about legged robots' efficient locomotion based on gait selection.~\cite{4Autonomous_gait_transition} builds a new quadruped robot with mechanisms enabling walking to galloping and applies CPGs to build the locomotion controller. They find that the robot would reach a lower CoT with a galloping gait instead of trotting when it moves at a high velocity.
To minimize energy consumption and determine a suitable switching strategy, some works~\cite{6Selecting_gaits, 7smit2017energetic} utilize the optimization method to determine the relevancy between gait and CoT. They focus on selecting the optimal gait for achieving efficient locomotion at a specific velocity and reveal the difference between the natural principles and the mathematical models. However, their conclusion still needs to improve when applied to a 3D-legged robot model since they ignore the quadruped robot's real-time stability control and gait transition.

To the best of our knowledge, few researchers consider combining the locomotion stability requirement with energy utilization simultaneously to achieve both stable and efficient locomotion for quadruped robots, while this trade-off mechanism happens on the natural quadrupeds at almost every locomotion moment. Therefore, this work evaluates a quadruped robot's locomotion stability and energy consumption on different terrains with various velocities. Then, a gait mapping is built from the robot's speed to gait patterns by considering different locomotion requirements. Finally, this paper accomplishes the quadruped robot's stable and efficient locomotion on different terrains through gait selection and transition.
\section{Methodology}
Our proposed strategy contains two parts: one is the transition mechanism, which includes the definition of different gait patterns and the switching among five distinct gaits; the other is the gait map built by the STB and CoT indexes' trade-off.

\subsection{Gait Patterns}
Quadruped animals lift their legs and touch down in a coordinated cyclic manner, known as gait, to achieve balance and efficient locomotion~\cite{ref-biomechanical}. Inspired by natural quadrupeds, we also define five gait patterns for quadruped robots and implement them on flat and slope terrains.

\begin{figure}
	\centering
	\includegraphics[width=\linewidth]{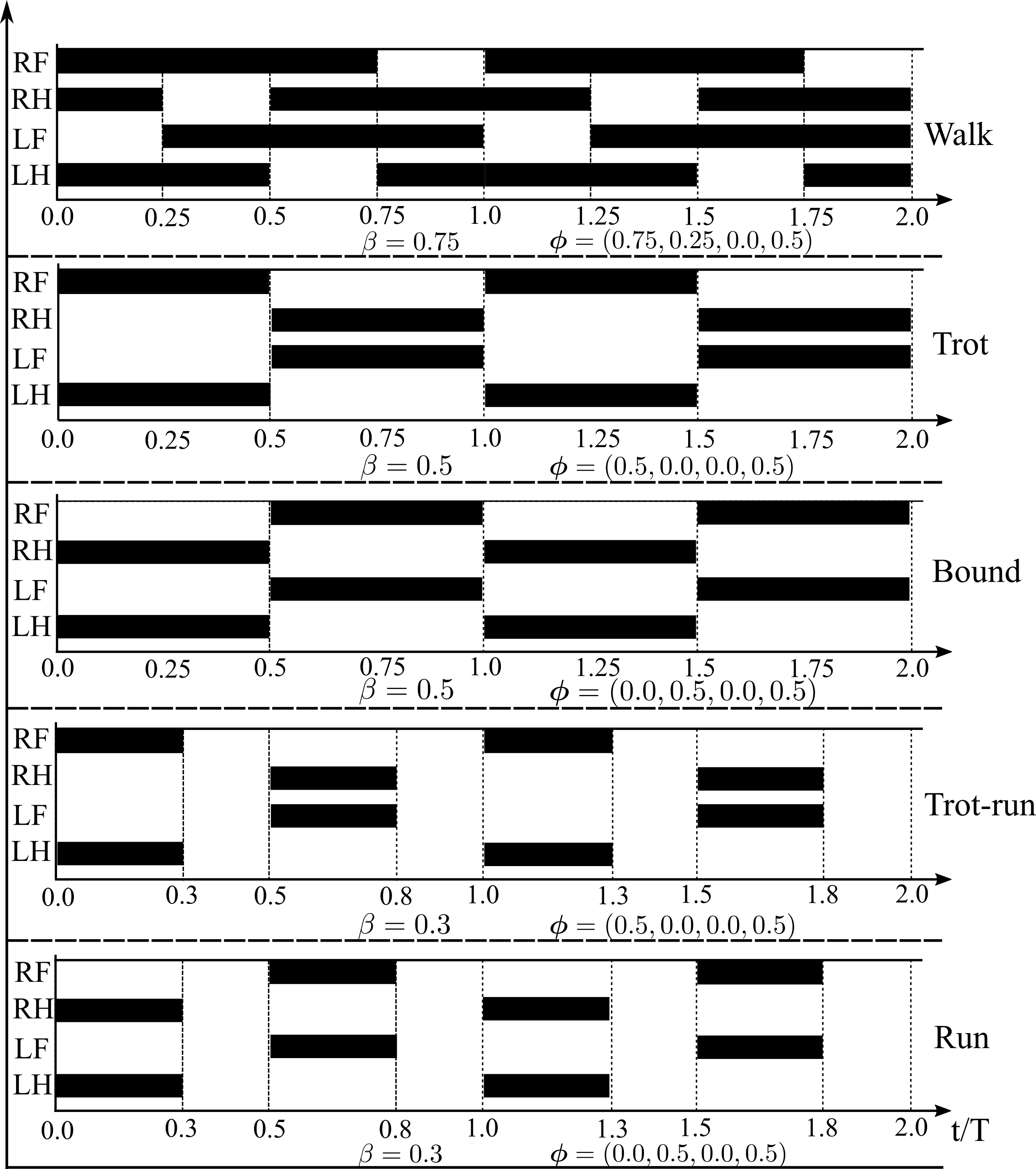}
	\caption{A phase diagram of five classical gait patterns for a quadruped robot. The black blocks represent the stance phase, while the blank ones represent the swing phase for each leg. T represents one stride period.}
	\label{gait_diagram}
 \vspace{-0.5cm}
\end{figure}

The gait patterns can be characterized as the representation of gait parameters, duty factor $\beta$ and the four legs' phase offsets $\boldsymbol{\phi}$ defined as $(\phi_{RF}, \phi_{RH}, \phi_{LF}, \phi_{LH})$ which represent the swing phase offset in the right front leg, right hind leg, left front leg and left hind leg, respectively. The duty factor $\beta$ represents the ratio of one leg stance duration to one stride period, and each leg's phase offset $\phi$ is defined as the lift-off (LO) timing in one stride period:
\begin{equation}
\begin{aligned}
& \beta = \frac{T_{st}}{T_{st}+T_{sw}} \\
&\phi = \frac{t_{LO}}{T_{st}+T_{sw}}
\end{aligned}
\end{equation}
where $T_{st}$ represents the stance duration and $T_{sw}$ represents the swing duration; the sum of them is a stride period. All five gait patterns used in this work are illustrated as a phase diagram shown in Fig.~\ref{gait_diagram}.

Walking is a static gait for its duty factor $\beta > 0.5$, which means at least three feet of contact with the ground at each moment. For trot and bound gait, their duty factor $\beta$ is 0.5, which represents keeping at least two contact points in one stride period without the flight phase (all feet are lifted off the ground). The difference between trot and bound gait is that the former's contact points are at a diagonal line while the other is at the front or hind of the robot. Therefore, trotting has less vibration than bounding when the robot starts moving. Trot-run and run gait are the upgrade modes for trotting and bounding with a flight phase. Since their duty factors $\beta = 0.3$ are less than 0.5, both can be considered dynamic gaits and perform better in energy efficiency than their original mode, as demonstrated in Sec.~\uppercase\expandafter{\romannumeral4}.

\subsection{Gait Transition}
The gait patterns are discrete for their different $\beta$ and $\boldsymbol{\phi}$, so it is hard to transform one gait into another instantaneously considering the physical limits. However, some gait patterns share the same parameters, making the transition between them much easier. For instance, trotting and bounding have the same $\beta$ and differ at $\boldsymbol{\phi}$. Therefore, we segment the continuous transition among all the gait patterns into four switching processes based on similarity, which are walk$\leftrightarrow$trot, trot$\leftrightarrow$bound, trot$\leftrightarrow$trot-run, and bound$\leftrightarrow$run. After that, the switching process between walk and trot can be built as follows:
\begin{equation}
	\begin{aligned}
	 &\beta(t) = \alpha_{0} \oplus \frac{1}{4T_{s}}t, t\in[0,T_{s}]\\
	 &\phi_{LF} = 0.0, \phi_{LH} = 0.5,\phi_{RF}= \beta(t), \phi_{RH} = \beta(t) \oplus 0.5\\
	 &trot \rightarrow walk:\oplus = +, \alpha_{0} = 0.5\\
	 &walk \rightarrow trot:\oplus = -, \alpha_{0} = 0.75
	\end{aligned}
\end{equation}
In the same way, another three switching processes can be characterized as follows, where trot $\leftrightarrow$ bound is:
\begin{equation}
\begin{aligned}
&\beta(t) = 0.5, t\in[0,T_{s}]\\
&\phi_{LF} = 0.0, \phi_{LH} = 0.5\\
&trot \rightarrow bound:\phi_{RF} = 0.5 - \frac{1}{2T_{s}}t, \phi_{RH} = \frac{1}{2T_{s}}t\\
&bound \rightarrow trot:\phi_{RF} = \frac{1}{2T_{s}}t, \phi_{RH} = 0.5-\frac{1}{2T_{s}}t
\end{aligned}
\end{equation}
trot $\leftrightarrow$ trot-run is:
\begin{equation}
\begin{aligned}
&\beta(t) = \alpha_{0} \oplus \frac{1}{5T_{s}}t, t\in[0,T_{s}]\\
&\phi_{LF} = 0.0, \phi_{LH} = 0.5,\phi_{RF}= 0.5, \phi_{RH} = 0.0\\
&trot \rightarrow trot-run:\oplus = -, \alpha_{0} = 0.5\\
&trot-run \rightarrow trot:\oplus = +, \alpha_{0} = 0.3
\end{aligned}
\end{equation}
bound $\leftrightarrow$ run is:
\begin{equation}
\begin{aligned}
&\beta(t) = \alpha_{0} \oplus \frac{1}{5T_{s}}t, t\in[0,T_{s}]\\
&\phi_{LF} = 0.0, \phi_{LH} = 0.5,\phi_{RF}= 0.0, \phi_{RH} = 0.5\\
&bound \rightarrow run:\oplus = -, \alpha_{0} = 0.5\\
&run \rightarrow bound:\oplus = +, \alpha_{0} = 0.3
\end{aligned}
\end{equation}
where $T_{s}$ is the transition time for each gait, and $\oplus$ is an operator representing $+$ or $-$ according to the transition. All the switching between two similar gaits can be accomplished during the transition time. Considering trot-run and run are both dynamic gaits with high instability, the direct transition between them will be denied, for it easily causes the robot to fall.
\begin{figure}[t]
	\centering
	\includegraphics[width=0.9\linewidth]{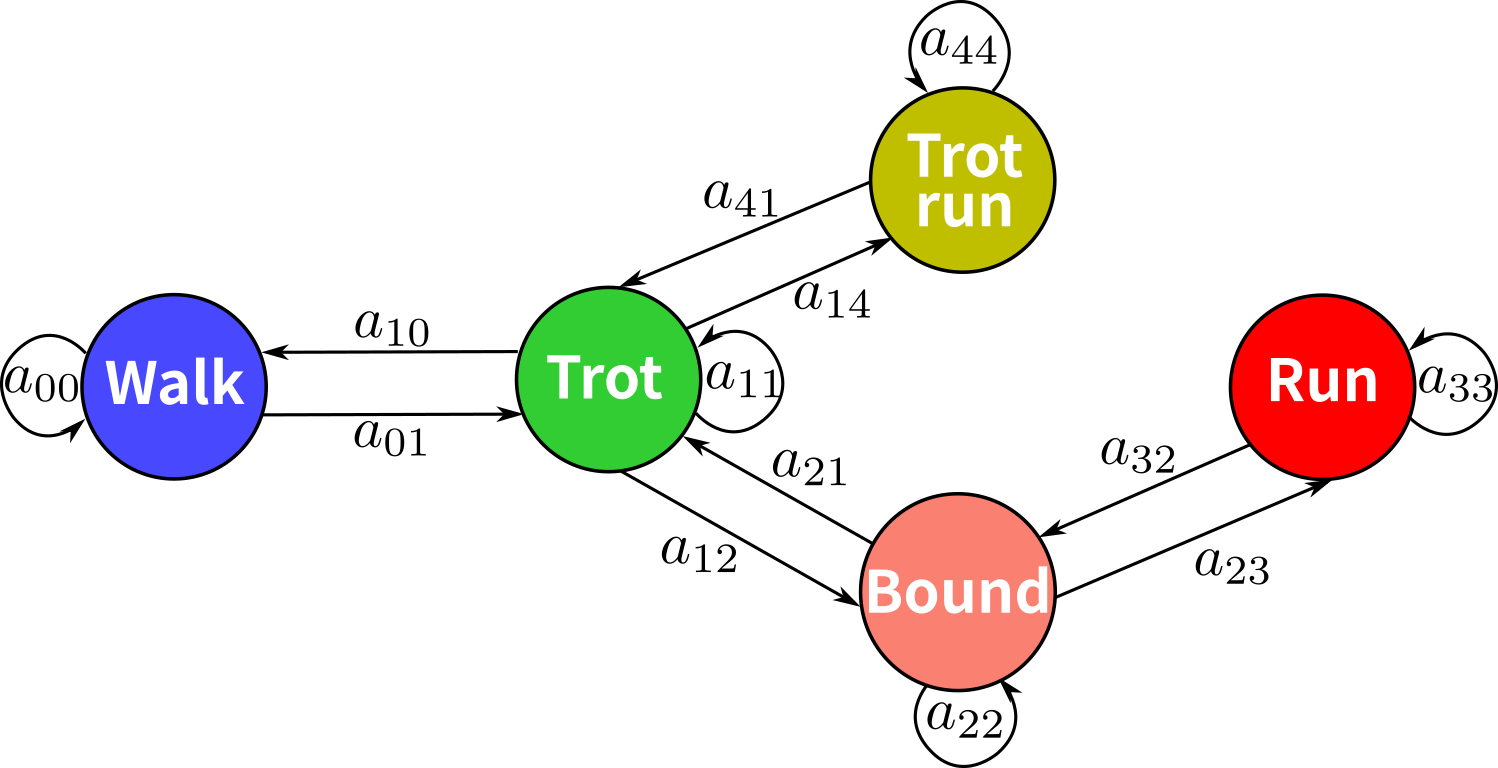}
	\caption{The graph for switching between gait patterns. $a_{ij}$ represents the switching action from state $i$ to state $j$.}
	\label{switch_action}
\end{figure}
\begin{table}[t]
    \footnotesize
    \renewcommand\arraystretch{1.1}
    \setlength{\tabcolsep}{1.5pt}
	\caption{Table of gaits transition order, E$_{i}$ represents the event of switching to the state $i$}
	\label{transition_table}
	\centering
		 
		\begin{tabular}{|c|c|c|c|c|c|}
			\hline
			\rule{0pt}{8pt}
			\multirow{2}{*}{\textbf{States}}&
			\multicolumn{5}{c|}{\textbf{Events}} \\
			\cline{2-6}
			\rule{0pt}{8pt}
			& \textbf{E$_{0}$} & \textbf{E$_{1}$} & \textbf{E$_{2}$} & \textbf{E$_{3}$}&\textbf{E$_{4}$}\\
			\hline
			\rule{0pt}{8pt}
			Walk & $a_{00}$ & $a_{01}$ & $a_{01}, a_{12
			}$ & $a_{01},a_{12}, a_{23}$ & $a_{01}, a_{14}$\\
			\hline
			\rule{0pt}{8pt}
			Trot & $a_{10}$ & $a_{11}$ & $a_{12}$ & $a_{12}, a_{23}$ & $a_{14}$\\
			\hline
			\rule{0pt}{8pt}
			Bound & $a_{21}, a_{10}$ & $a_{21}$ & $a_{22}$ & $a_{23}$ & $a_{21}, a_{14}$\\
			\hline
			\rule{0pt}{8pt}
			Run & $a_{32},a_{21}, a_{10}$ & $a_{32}, a_{21}$ & $a_{32}$ & $a_{33}$ & $a_{32},a_{21},a_{14}$\\
			\hline
			\rule{0pt}{8pt}
			Trot-run & $a_{41}, a_{10}$ & $a_{41}$ & $a_{41}, a_{12}$ & $a_{41}, a_{12}, a_{23}$ & $a_{44}$\\
			\hline
		\end{tabular}
\end{table}

Locomotion controllers for legged robots need to manage the continuous dynamics of the system and the discrete gait transition. For this reason, we design a graph to define the switching action between gaits, as shown in Fig.~\ref{switch_action}. Furthermore, a finite state machine (FSM) is developed to set the switching order among all the gait patterns, and we define the related state numbers for different gaits: walking, trotting, bounding, running, and trot-running are represented as {0, 1, 2, 3, 4}, respectively. The events that realize transitions between states are achieved through no more than three actions, as shown in Tab.~\ref{transition_table}.

\subsection{Gait Selection}
Achieving efficient and stable locomotion for quadruped robots depends on the gait transition and determining an appropriate gait for different terrains. 
Therefore, selecting a suitable gait to adapt to various terrains is necessary and valuable.

Firstly, the energy consumption during locomotion needs to be figured out. The total power is the sum of the mechanical motor power and the motor copper losses of all joints. Since our experiments are conducted on the quad\_sdk~\cite{quad_sdk}, a quadruped robot simulator based on ROS and Gazebo, the robot will not produce heat losses and cannot store excess electrical energy in batteries or capacitors. If all motors create negative power, this power should be considered dissipating in the system. Then, the consumption energy $W$ can be integrated over a full stride as follows~\cite{6Selecting_gaits}:
\begin{equation}
	W = \int_{0}^{t_{f}} (\sum_{i=1}^{n} max(u_{i} \cdot \omega_{i}, 0)) dt
\end{equation}

To compare energetic economy across different terrains with various velocities, the cost of transport (CoT) is introduced:
\begin{equation}
	\text{CoT} = \frac{W}{mg\Delta s}
\end{equation}
where $t_{f}$ represents one stride period, $n=12$ is the motor numbers of the robot, $u_{i}$ and $\omega_{i}$ represent the torque and the angular velocity of each motor respectively, $m$ is the total mass of the robot and $\Delta s$ represents the travel distance in one stride period.  

Furthermore, the robot should remain stable enough to guarantee basic locomotion in different environments.  Hence, we build an index to express the robot's stability under locomotion called STB, which is a variant of the $J_{3}$ in~\cite{5Automatic_gait_selection}.
\begin{equation}
	\text{STB} = w_{1}(|v_{bn}/v_{b}|) + w_{2}|\theta_{b} - \theta_{t}|+w_{3}|\phi_{b}|+w_{4}(|\dot{\theta}_{b}|+|\dot{\phi}_{b}|))
\end{equation}
where $v_{bn}$ represents the normal velocity of the robot, $v_{b}$ is the robot's actual velocity, $\theta_{b}$ and $\phi_{b}$ represent the body's pitch and roll angle, respectively. $\theta_{t}$ is the inclination of the terrain and $\dot{\theta}_{b}$, $\dot{\phi}_{b}$ are the attitude change rate. STB integrates the robot's attitude variation and vertical vibration to represent body stability. Apparently, the unstable locomotion is mainly caused by the robot's attitude, so the weight of attitude terms $w_{2}, w_{3}$ are set as 1.0,  $w_{1}$ and $w_{4}$ are set as 0.7 and 0.3. 

Both CoT and STB will be affected by gait patterns $\boldsymbol{\Lambda}$ and should be considered simultaneously. Although the CoT and STB have different dimensions, they share a similar scale during normal locomotion (from 0 to 1.4, which can be seen in Fig.~\ref{gait-curve}). So, the two indexes are integrated into one total index $J_{e}(\boldsymbol{\Lambda})$:
\begin{equation}
	J_{e}(\boldsymbol{\Lambda}) = c \cdot \text{STB}+(1-c)\cdot \text{CoT}
	\label{J_e}
\end{equation} 
where $c$ is the ratio of STB in the evaluation equation and can be altered to achieve different goals.

After that, the most suitable gait $\boldsymbol{\Lambda}^{*}$ for different terrains can be acquired by minimizing the $J_{e}$ while satisfying the robot's dynamics and other constraints. Therefore, the optimization formulation can be constructed as follows:
\begin{subequations}
	\begin{align}
	&min \quad (\sum_{i=1}^N J_{e}(\boldsymbol{\Lambda})) / N \label{Za}\\
s.t. \quad \quad \quad	
		&m \ddot{\boldsymbol{r}} = m\boldsymbol{g}+\sum_{l=1}^{4}\boldsymbol{f}_{l}\notag\\
&I\boldsymbol{\dot{\omega}}+\boldsymbol{\omega}\times(I\omega)=\sum_{l=1}^{4}\boldsymbol{f}_{l}\times (\boldsymbol{r} - \boldsymbol{p}_{l}) \label{Zc}	     \\
& -\mu \boldsymbol{f}_{l}^{n} \leq \boldsymbol{f}_{l}^{t} \leq \mu  \boldsymbol{f}_{l}^{n} \quad  \quad stance \quad leg  \label{Zd}  \\
& \boldsymbol{I}\boldsymbol{f}_{l} = \boldsymbol{0} \quad \quad \quad \quad \quad \quad \quad swing \quad leg \label{Ze}
	\end{align}
	\label{min_Je}
\end{subequations}
where Eq.~\eqref{Za} is the optimization goal, and $N$ is the pre-defined strides' count. Eq.~\eqref{Zc} represents the dynamics constraints, where $\boldsymbol{r} \in \mathbb{R}^{3} $, $\boldsymbol{\omega} \in \mathbb{R}^{3} $ are the Center of Mass (CoM) position and angular velocity, and $\boldsymbol{p}_{l} \in \mathbb{R}^{3} $,  $\boldsymbol{f}_{l} \in \mathbb{R}^{3}  $represent the foot location and contact forces of leg $l$ expressed in the world frame. Eq.~\eqref{Zd} is the friction cone that prevents the slipping of the robot's feet, where $\mu$ is the friction coefficient, $\boldsymbol{f}_{l}^{n}$ represents the normal component of the contact force, and $\boldsymbol{f}_{l}^{t}$ is the tangential component. $\boldsymbol{I} \in \mathbb{R}^{3\times3}$ is the identity matrix and $\boldsymbol{0} \in \mathbb{R}^{3}$.

Extensive experiments are conducted to find the optimal gait on each terrain and ratio $c$ with various velocities. Then, we can transform these data into a velocity-gait map on each terrain by calculating Eq.~\eqref{Za}. Finally, the gait selection is completed by querying the map on any related terrains.


\section{Experimental Results and Discussion}
The main focus of this work is to build a multi-gait strategy for stable and efficient quadruped robot locomotion on various terrains. Therefore, all experiments in this work are conducted to support our key claims, which are: (i)~ realizing the smooth switching among five distinct gaits and achieving highly dynamic gait motion; (ii)~constructing a mapping mechanism from energy consumption, velocities, and stability to the gait patterns on different terrains through the evaluated information fusion; (iii)~building an evaluation mechanism based on energy efficiency and locomotion stability, and demonstrating the practical applicability of the gait selection theory from biology to quadrupedal robots. 
\begin{figure}[thpb]
	\centering
	\subfigure[Trotting $\rightarrow$ Walking $\rightarrow$ Trotting $\rightarrow$ Trot-running]{\includegraphics[width=0.80\linewidth]{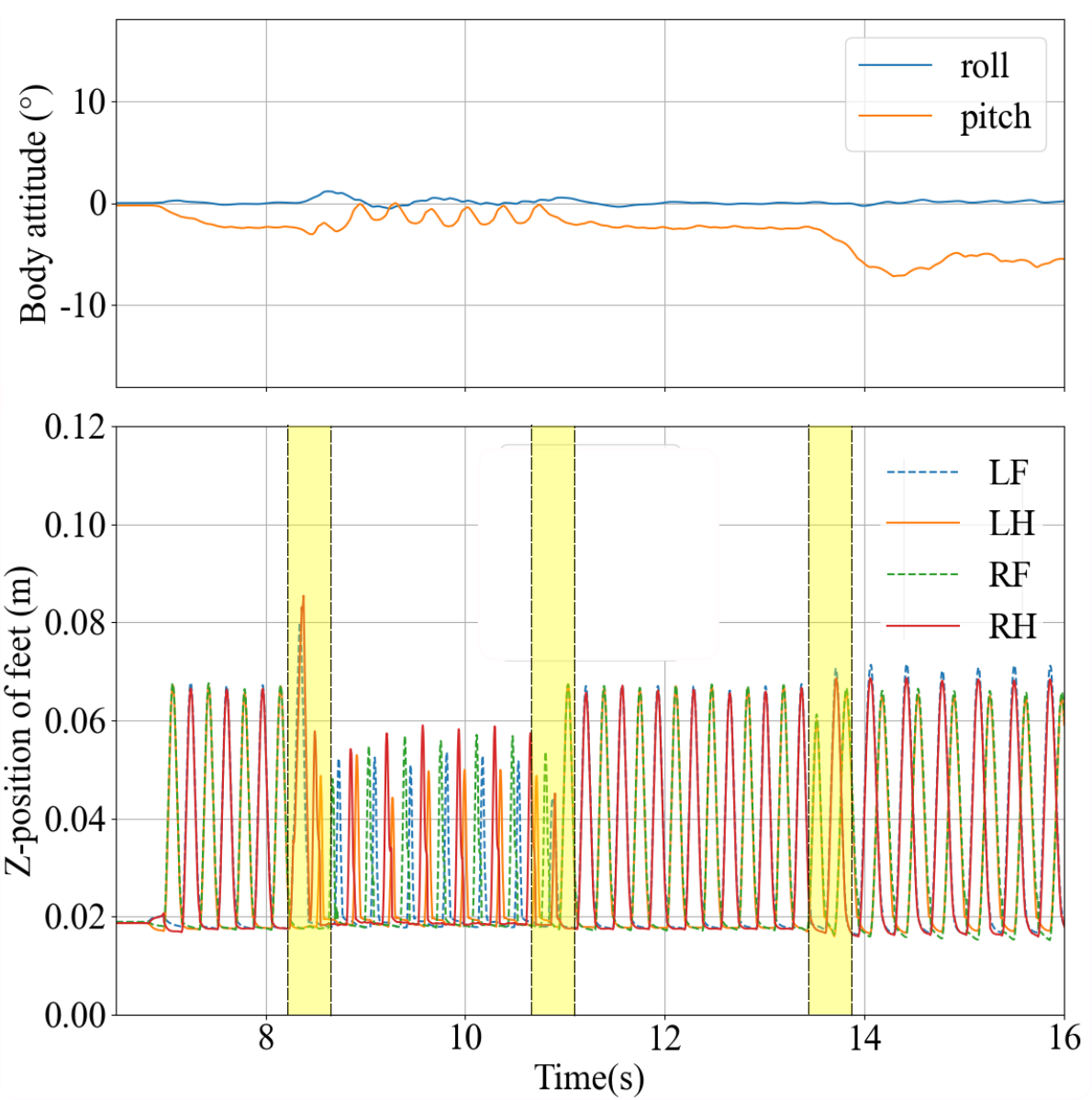}}
	\\
	\subfigure[Trotting $\rightarrow$ Bounding $\rightarrow$ Trotting $\rightarrow$ Running]{\includegraphics[width=0.80\linewidth]{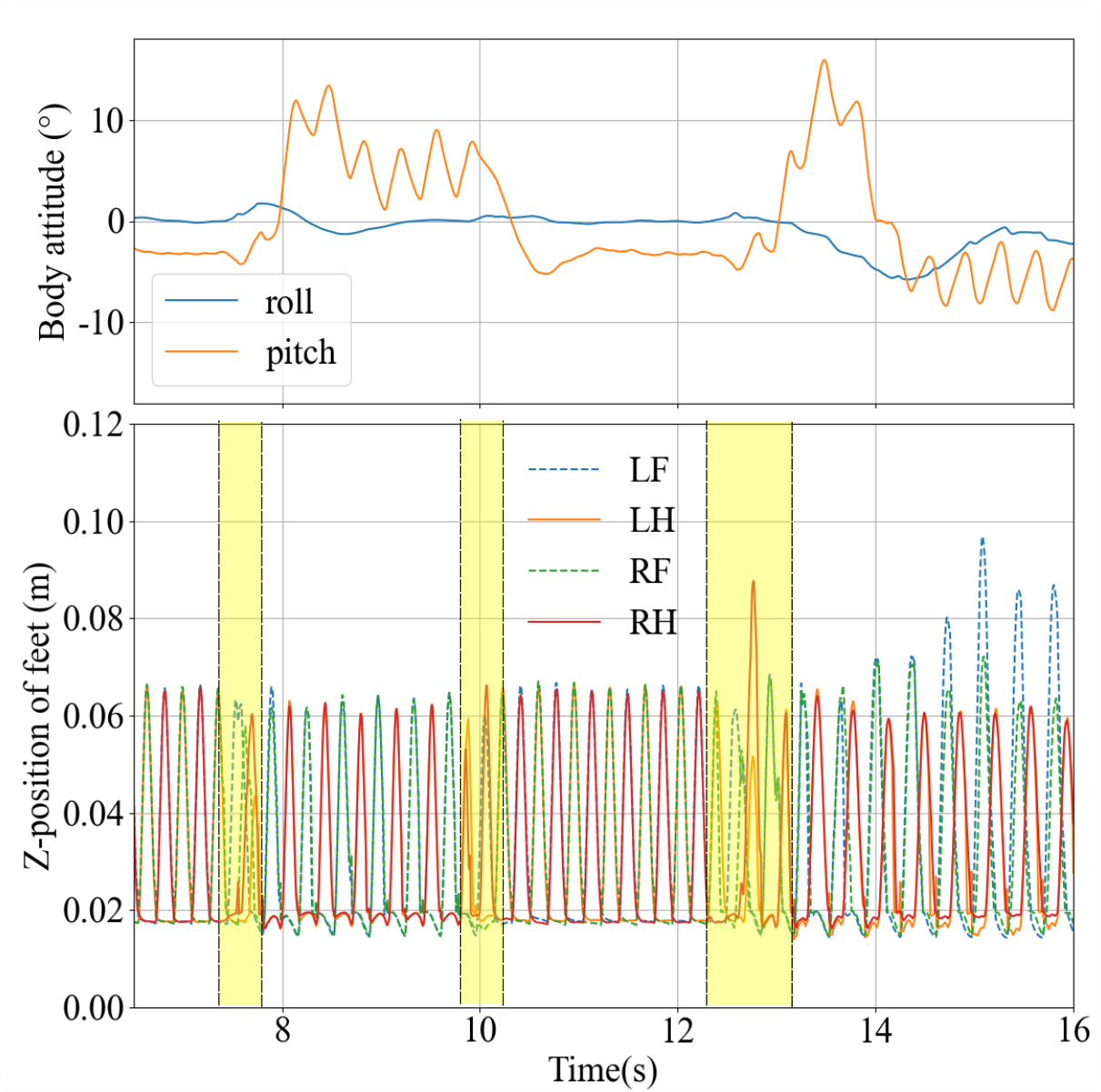}}
	\caption{The feet's height variation in transition processes. The upper plot of each subfigure shows the roll and pitch angle of the robot's body, while the bottom one shows the feet's position in the z-direction, and the yellow parts represent the transition process.}
	\label{gait-transition}
 \vspace{-0.3cm}
\end{figure}
\subsection{Gait Transition Experiment}
The first experiment is to evaluate the performance of the proposed transition method and its outcomes to support the claim that we realize the smooth transition among five distinct gait patterns and achieve the highly dynamic gait locomotion ( running gait and trot-running gait). 

The proposed gait transition method is tested on flat terrain with transition duration $T_{s}$ set to 0.5\,s, and the results are shown in Fig.~\ref{gait-transition}. Fig.~\ref{gait-transition}~(a) shows the transition from trotting to walking and trot-running with the robot's speed is 1.2\,m/s. At the beginning of Fig.~\ref{gait-transition}~(a), the RF, HL curves are overlapped, and the same for the LF, RH curves, which represent the robot using the trotting gait. After that, the robot switches its gait to walking (from 8.5 s to 10.5 s) and to trot-running (after 14 s). In the walking gait part, there are four distinct peaks, representing that only one foot is in the air during each moment. In trot-running gait, the LF and RH curves overlap while the four curves intersect at the same point whose ordinate is higher than 0.02 m (the ground clearance), representing all feet in the air (the whole robot is in the flight phase). This result shows that the robot can maintain balance and attitude in a narrow area under trot-running gait, which verifies our first claim. 

Fig.~\ref{gait-transition}~(b) shows the transition from trotting to bounding and running at 1.9\,m/s. The RF, LF curves are overlapped and the same for the LH, RH curves, which represent the robot using the bounding gait from 8 s to 10 s. After that, the robot switches its gait to trotting (from 10.5 s to 12s) and running (after 13 s). In the running gait part, the LF and RF curves still overlap while the four curves intersect at a high point, representing the whole robot in the flight phase. Additionally, the pitch angle curve represents that the robot has periodical vibration while still maintaining balance when taking the running gait. This illustrates that our method can achieve highly dynamic gait locomotion.

Beyond that, we can find that bounding and running cause more body vibration compared to other gaits. Apparently, trotting performs the most stable among the three gaits according to the variation of body attitude in the upper plot of Fig.~\ref{gait-transition}~(a), which is due to its diagonal stepping without flying phase. 
 \begin{figure}[thpb]
 	\centering
 	\subfigure[The STB and CoT values for different gait patterns with various velocities on flat terrain.]{\includegraphics[width=0.95\linewidth]{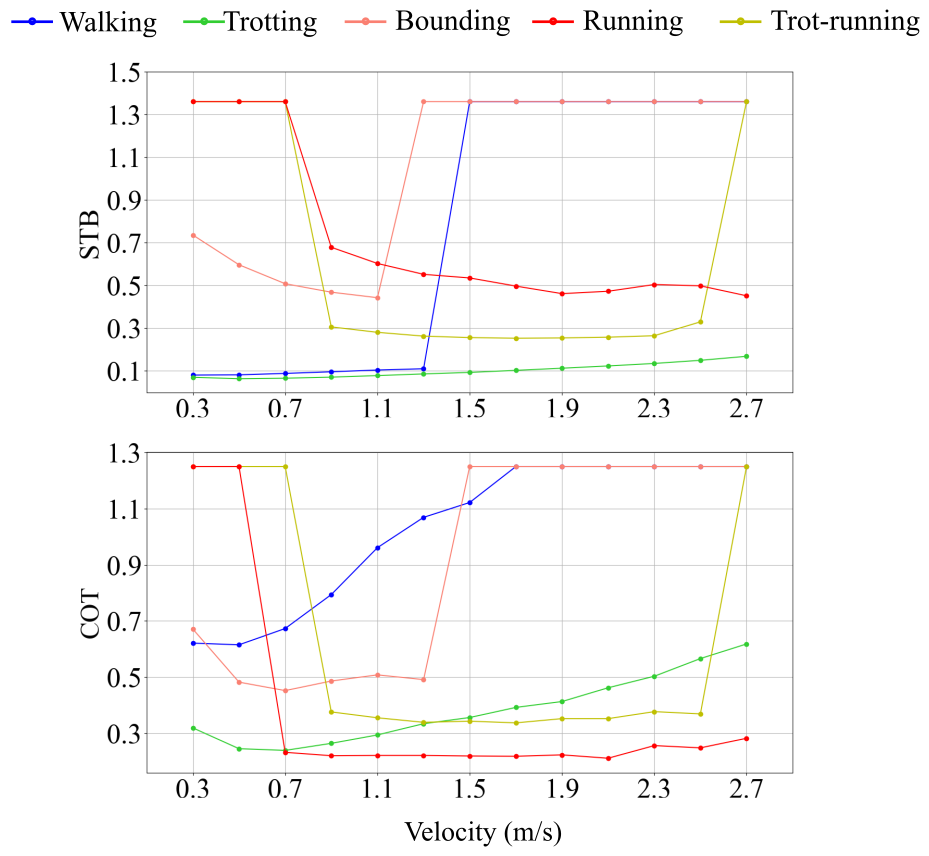}}
 	\\
 	\subfigure[The STB and CoT values for different gait patterns with various velocities on a 12\,$^{\circ}$ slope.]{\includegraphics[width=0.95\linewidth]{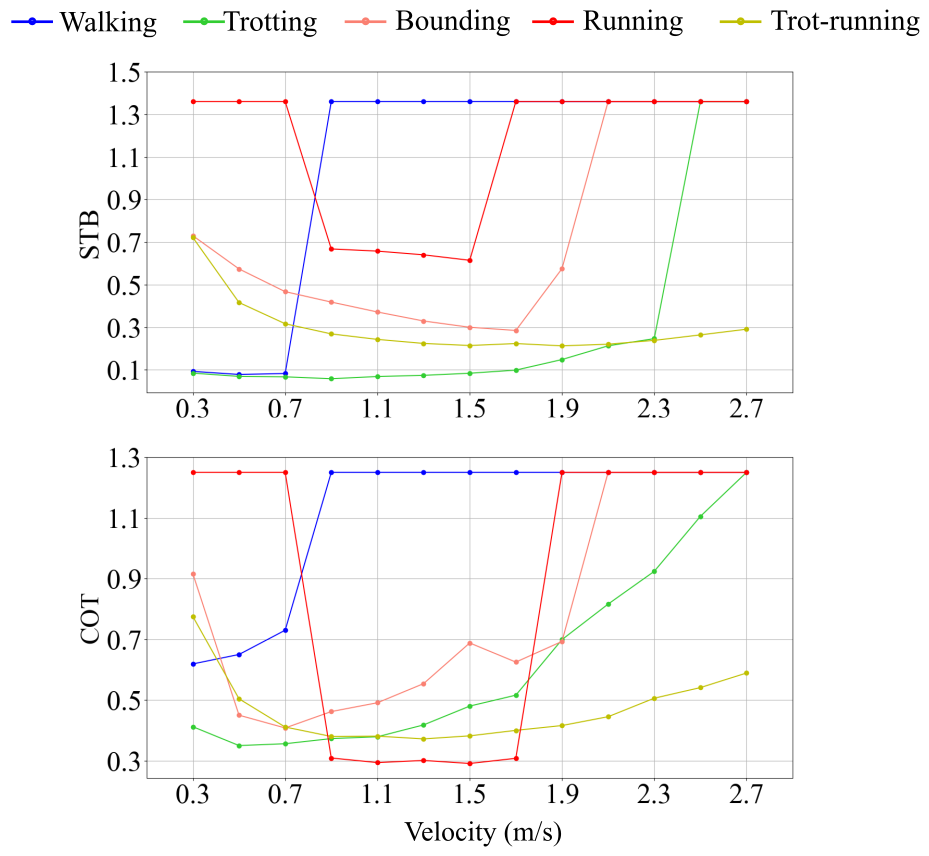}}
 	\caption{The index value of different gait patterns on two terrains.}
 	\label{gait-curve}
  \vspace{-0.3cm}
 \end{figure}
\subsection{Velocity-gait Mapping}
The second experiment explains the construction of the velocity-gait mapping and illustrates that our approach is capable of achieving the most suitable gait for quadruped robots on different terrains with velocities varying. To form the velocity-gait mapping, we collect the CoT and STB indexes with various velocities from 0.3\,m/s to 2.7\,m/s on flat and slope terrains, respectively. Each test with a fixed velocity on one terrain is repeated at least five times, and the curves of the average results are plotted in Fig.~\ref{gait-curve}.
 \begin{figure}[thpb]
	\centering
	\subfigure[The velocity-gait mapping on a flat.]{\includegraphics[width=0.9\linewidth]{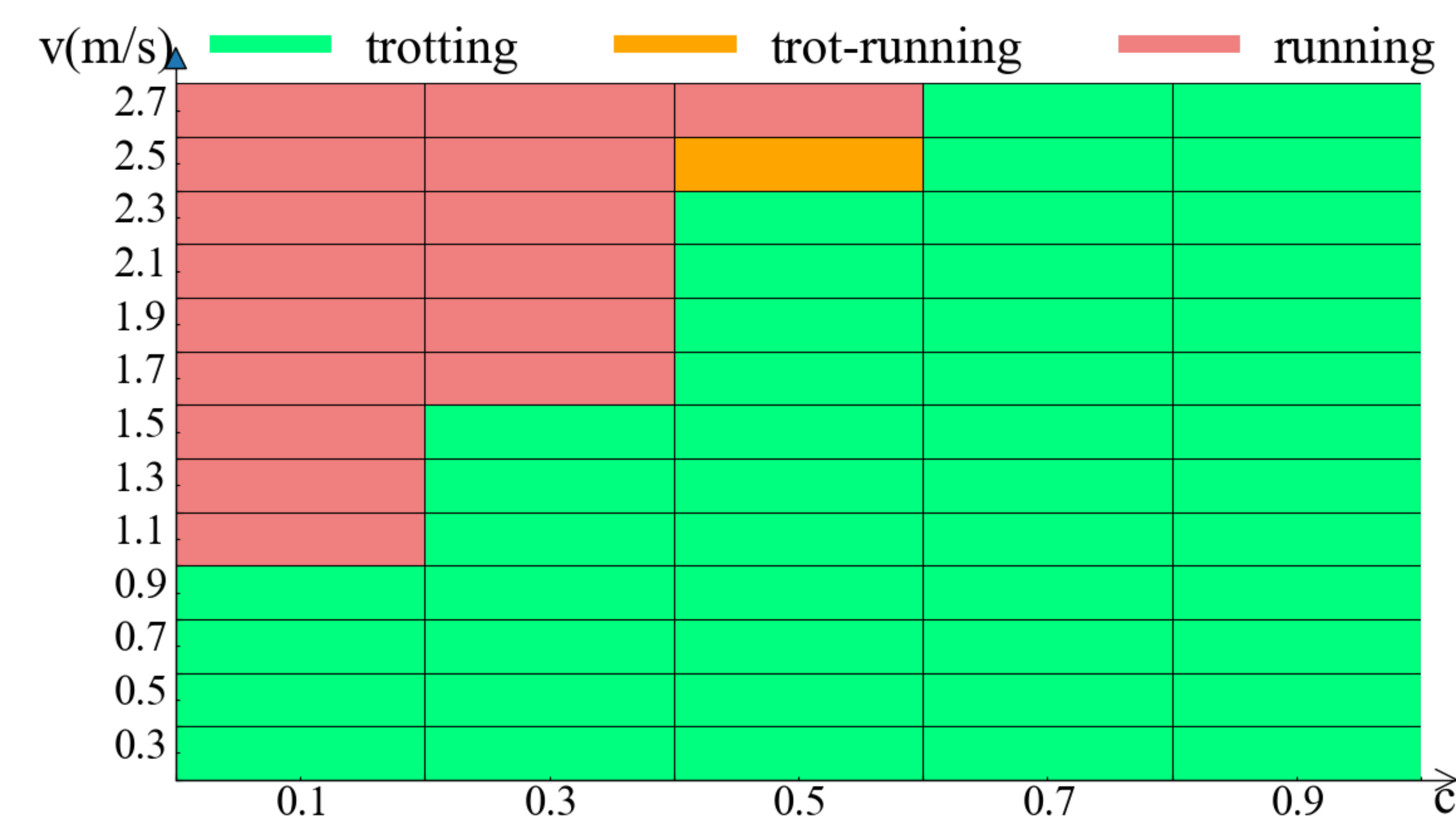}}
 \vspace{-0.2cm}
	\\
	\subfigure[The velocity-gait mapping on a 12\,$^{\circ}$ slope.]{\includegraphics[width=0.9\linewidth]{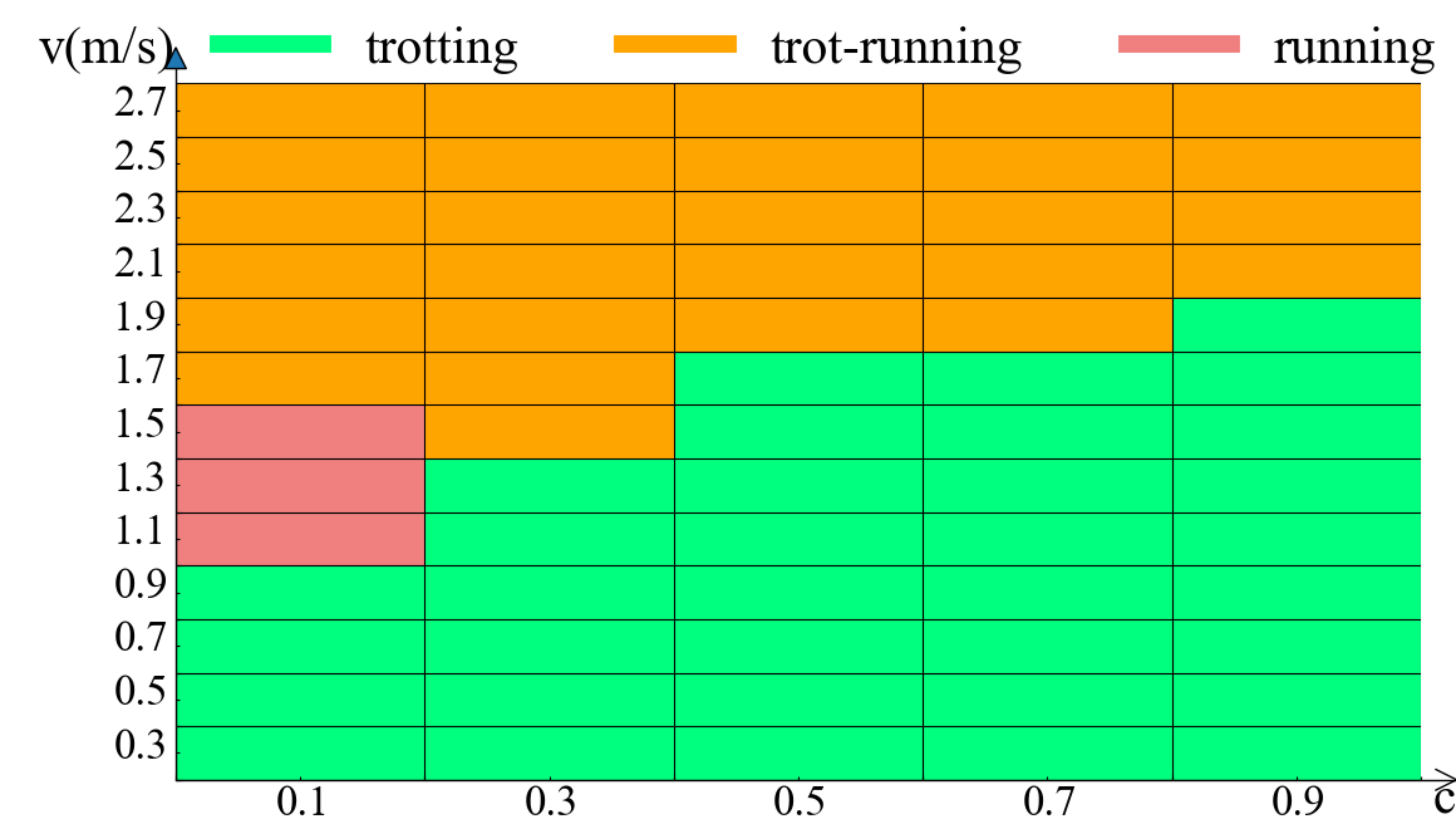}}
 \vspace{-0.2cm}
	\caption{The gait selection mapping with various velocities on two terrains. $c$ represents the coefficient of STB in $J_{e}$}
	\label{gait-map}
 \vspace{-0.5cm}
\end{figure}
\vspace{0.5cm}

Fig.~\ref{gait-curve}~(a) illustrates that trotting has the lowest STB on flat terrain while running shows higher energy efficiency with lower CoT when the robot's velocity is faster than 0.7\,m/s. Fig.~\ref{gait-curve}~(b) demonstrates that trotting is the most stable gait with a velocity less than 2.3\,m/s, and trot-running saves more energy when the robot moves at high speed on a slope. Besides, when the robot takes locomotion on a terrain, it may inevitably fall at some speed because of the controller's limitation. If the robot falls down at each test, the correct CoT and STB values can not be collected. Therefore, the CoT and STB upper bounds are set as 1.25 and 1.36, respectively, near the maximum values in all normal tests. Furthermore, it is unreasonable to decide a gait for different velocities with only one index value. Therefore, combining two indexes according to Eq.~\eqref{J_e} and Eq.~\eqref{min_Je} would be necessary, then we transform the merged data into a mapping from locomotion velocities to gait patterns. The results are shown in Fig.~\ref{gait-map}.
\begin{figure*}[thpb]
	\centering
	\subfigure[The illustration of the robot moving on a flat-slope terrain.]{\includegraphics[width=0.95\linewidth]{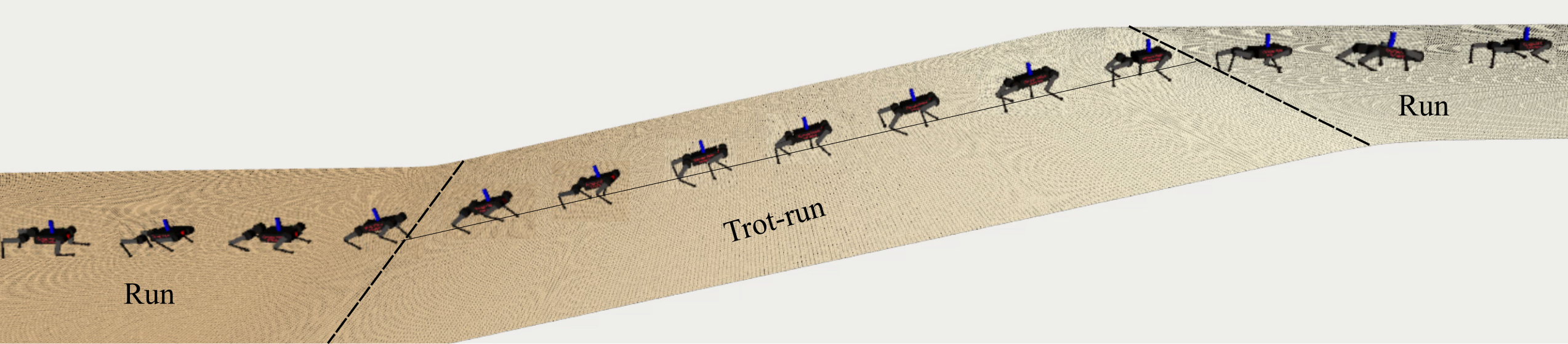}}
 \vspace{-0.2cm}
	\\
 \subfigure[The illustration of the robot moving on a continuous slope terrain.]{\includegraphics[width=0.95\linewidth]{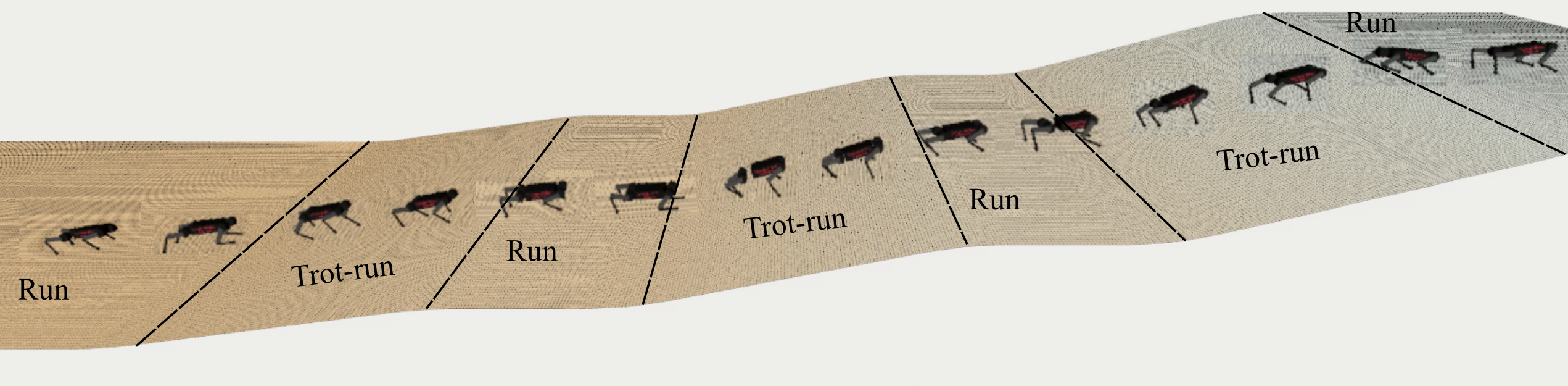}}
  \vspace{-0.2cm}
	\\
 \subfigure[The illustration of the robot moving on a up-down slope terrain.]{\includegraphics[width=0.95\linewidth]{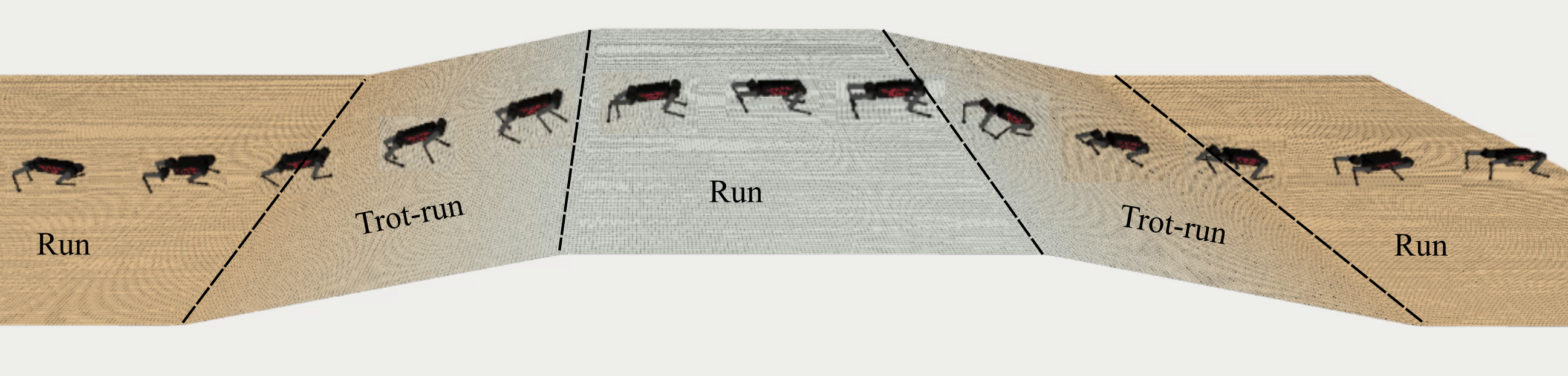}}
	\caption{The experiments of the multi-gait strategy using running gait on the flat and trot-running gait on the slope.}
	\label{experiment}
 \vspace{-0.3cm}
\end{figure*}

On a flat, trotting is the most proper gait on each $J_{e}$ index when the robot is at low speeds (up until 0.9\,m/s). However, if we consider the energy efficiency (c is no more than 0.3) at middle or high speeds (more than 1.1\,m/s). The model favors a running gait, which has been demonstrated by many research~\cite{4Autonomous_gait_transition, 6Selecting_gaits, ref-hoyt1981gait} as the best gait for saving energy in the natural quadrupeds with high speeds. Fig.~\ref{gait-map}~(b) shows that trot-running performs better on any $J_{e}$ index when the robot takes high-speed locomotion on a slope. The reason is that the trot-running gait can use diagonal stepping to achieve more stable locomotion than running and reduce the body's swing on pitch direction, especially for a slope. At the same time, trot-running saves more energy by adding the flying phase compared to trotting. Therefore, trotting and trot-running are the most suitable for locomotion on slope terrain.
Additionally, the result shown in Fig.~\ref{gait-map}~(b) coincides with the paper~\cite{5Automatic_gait_selection}, whose results also illustrate that trot-running is the optimal gait at high speeds (more than 1.7\,m/s) to maintain the robot's locomotion stability in most situations. In a word, trotting is the best gait when the robot is at low speeds, whether on a flat or slope. Running gait is more suitable for saving energy when moving on a flat at high speed, while trot-running is better on a slope.

\subsection{Multi-gait Strategy Experiment}
The last experiment is conducted to verify the efficiency of the proposed multi-gait strategy and demonstrate the practical applicability of the gait selection theory from biology to quadrupedal robots. The experimental process and related gaits are shown in Fig.~\ref{experiment}~(a). The experimental setup is as follows: the robot receives a speed command, which forces it to move forward; our strategy will select the proper gait guided by the gait mapping on the current terrain and accomplish the gait transition until it reaches another terrain. At the same time, we introduce other methods, including two fixed gait methods which take the same gait for each test,  and one multi-gait method which determines the optimal gait according to the velocity and does not switch gait during one test (Wang's method~\cite{5Automatic_gait_selection}) as baselines, then record the corresponding data for comparison. 
For our strategy, we take the stability coefficient $c$ from 0.1 to 0.9 in the gait mapping when the robot moves on the whole terrain; each method is tested and repeated 30 times on a random velocity from 0.3\,m/s to 2.7\,m/s with the average at 1.481\,m/s, the results are shown in Tab. ~\ref{table_1_3}.
\begin{figure*}[thpb]
	\centering
	\includegraphics[width=0.95\linewidth]{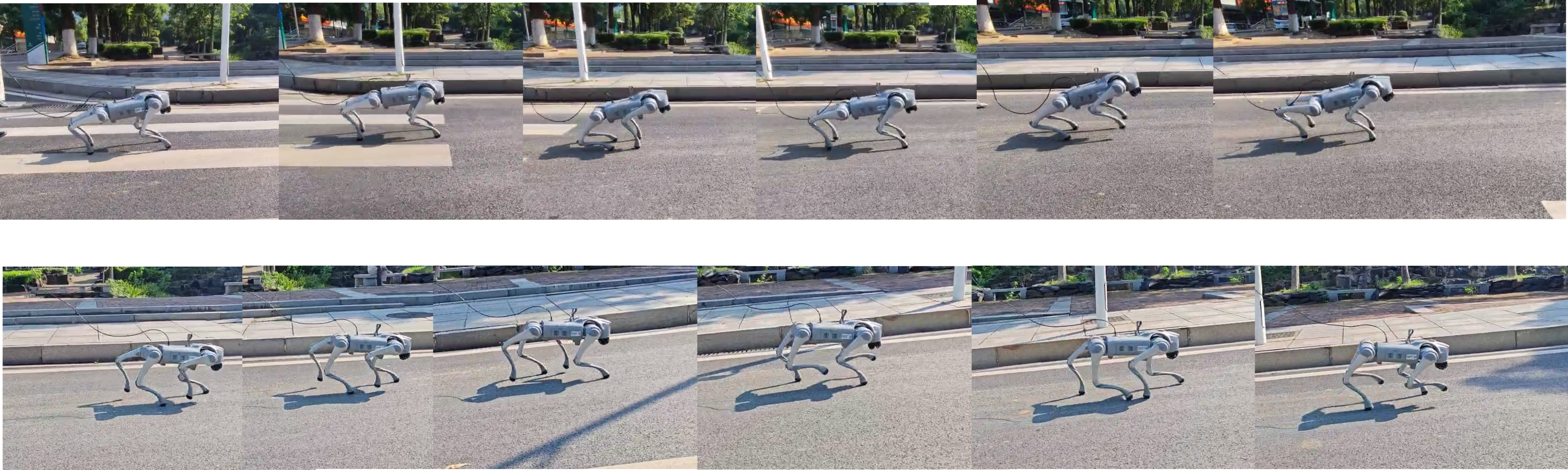}
    \caption{Go2 robot adopts the proposed strategy through flat-slope terrain. The top picture shows a Bound-run gait on flat ground and the bottom picture shows a Trot-run gait on a slope}
	\label{real_experiment}
 \vspace{-0.3cm}
\end{figure*}
\begin{figure}
	\centering
	\includegraphics[width=\linewidth]{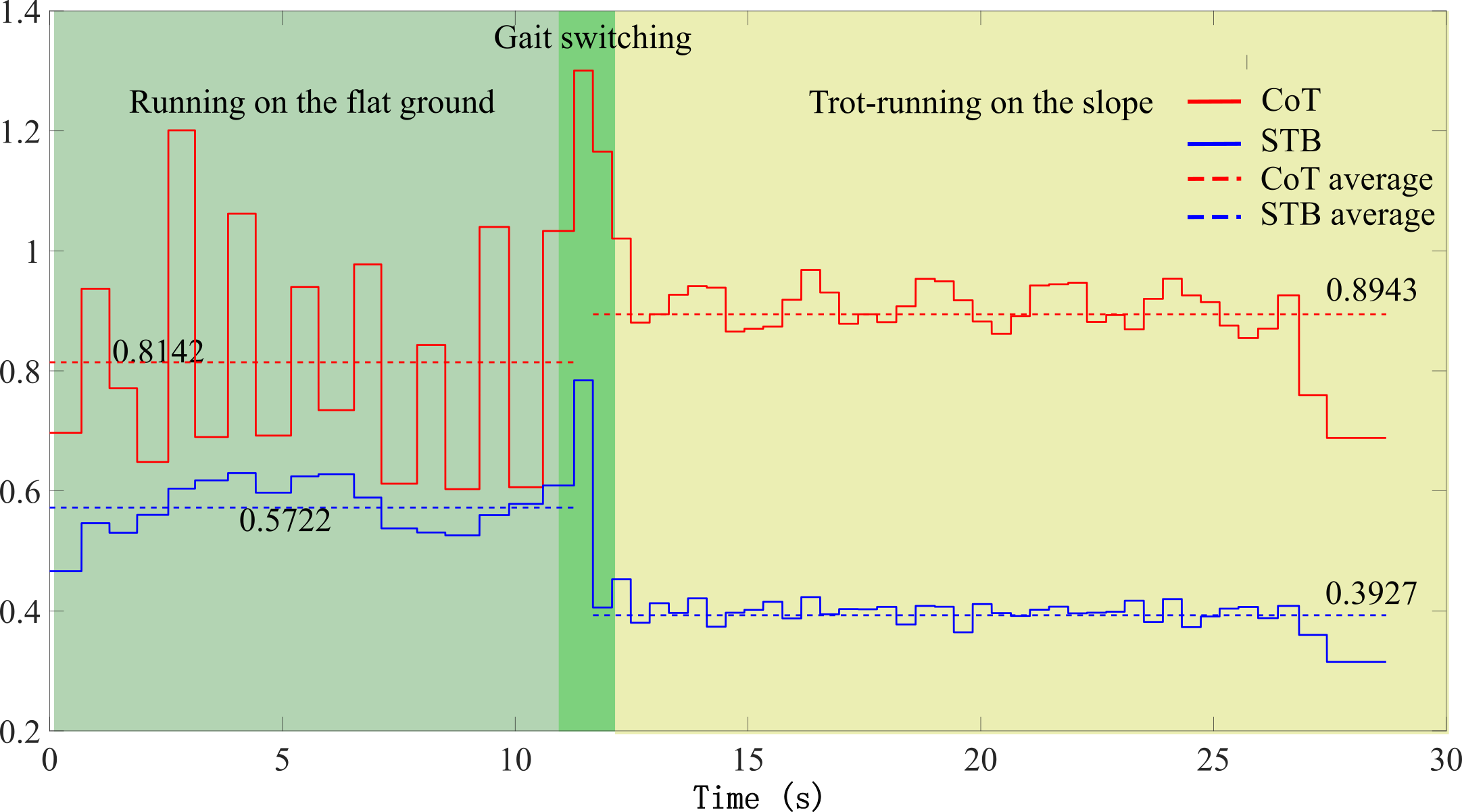}
	\caption{Real-world test data for multi-gait strategies on flat-slope terrain}
	\label{real_cot}
 \vspace{-0.5cm}
\end{figure}
\begin{table}[t]
\footnotesize
\setlength{\tabcolsep}{7pt}
\renewcommand{\arraystretch}{1.1}
	\caption{Comparison between different methods with the multi-gait strategy at a random velocity.} 
 \vspace{-0.2cm}
	\label{table_1_3}
	\centering
	 \begin{tabular}{lccc} 
		\toprule[1.0pt] 
		Gait strategy& CoT & STB & Success ratio\\
		\midrule 
		trotting & $0.383$ & $0.145$  & 29\,/30 \\
		trot-running& $0.532$ & $0.512$  & \color{red}23\,/30 \\
		Wang's method~\cite{5Automatic_gait_selection} & $0.331$  & $0.220$  & 29\,/30 \\
		Ours c1 ($c=0.1$) & $\boldsymbol{0.299}$  & $0.355$  & 29\,/30 \\
		Ours c3 ($c=0.3$) & $0.316$  & $0.283$  & 29\,/30 \\
		Ours c5 ($c=0.5$) & \color{orange}$0.330$  & \color{orange}$0.118$  & \color{orange}30\,/30 \\
		Ours c7 ($c=0.7$) & $0.340$  & $0.117$  & 30\,/30 \\
		Ours c9 ($c=0.9$) & $0.344$  & $\boldsymbol{0.109}$  & 30\,/30 \\
		\bottomrule 
	\end{tabular}
 \vspace{-0.5cm}
\end{table}

All the results are calculated by introducing the failed effect, which means if the robot falls on the test ( can not accomplish the entire process), the CoT and STB values are set as the maximum, 1.25 and 1.36, respectively. According to the results, our strategy ($c= 0.1$) shows the lowest energy consumption on the whole terrain among all methods, with only one failure. This strategy makes the robot run on flat terrain at high speed to improve energy utilization while trot-running on a slope to maintain stability. The trot-running method shows only a 76.7\,\% (23\,/30) success rate for its unstable locomotion on the flat at low velocities, as shown in Fig.~\ref{gait-curve}. Compared with other methods, our strategies with $c = 0.5, 0.7, 0.9$ reach a similar STB at the lowest level, and their success rates are 100\,\%, meaning our strategies can achieve the most stable locomotion on the whole terrain. At the same time, the proposed strategy c5 shares a similar CoT with Wang's method while reaching a lower STB than the latter. Therefore, our strategy c5 is the most compromised method to achieve stable and efficient locomotion among all the methods. Those results indicate that our multi-gait strategy enables the robot to achieve efficient and stable locomotion in the specified environment by selecting the proper gait and completing the transition. Additionally, the proposed method is also tested on some more complex terrains, which are the continuous slope (the inclinations are 8$^\circ$, 12$^\circ$ and 18$^\circ$) and the up-down slope (both the inclinations are 12$^\circ$). The results are shown in Fig.~\ref{experiment}~(b) and Fig.~\ref{experiment}~(c).

In the simulation test, the proposed multi-gait motion strategy achieves the desired motion effect. In order to verify that the multi-gait strategy can also be useful in real life, this paper carries out an outdoor experiment based on a Unitree go2 robotic dog, adopting the c1 strategy with a movement speed of 1.7 m/s, and the test environment is a flat-slope terrain, and the movement process is shown in Fig. \ref{real_experiment}. It can be seen that the multi-gait locomotion strategy makes the robot dog adopt running gait on flat ground and trot-run gait on slope, which is in line with the c1 strategy, and also agrees with the effect achieved in the simulation experiment. The motion data of the go2 robot was collected and the values of CoT and STB were plotted as shown in Fig. \ref{real_cot}. It can be seen that, after adopting the multi-gait motion strategy, although the switch from flat to sloping terrain makes the energy consumption of the robot dog rise slightly (overcoming gravity to do work on the slope inevitably leads to a rise in energy consumption), it significantly reduces the value of the STB and ensures the stability of the robot dog's motion on the slope, which proves that the proposed multi-gait locomotion strategy can be applied in practice and has the ability to balance the energy efficiency and stability of locomotion.

\section{Conclusion}
In this paper, we present a locomotion strategy for quadruped robots. It includes two parts: the gait selection, built by merging CoT and STB indexes at various speeds on different terrains, and the gait transition mechanism, which relies on FSM and the linear transformation of gait parameters. Our experiments on quadruped robots demonstrate that the proposed multi-gait locomotion strategy enables the robot to choose the most suitable gait on different terrains and achieve gait transition in real-time, whether the goal is to realize the more stable, the more efficient, or the faster locomotion.    

During the gait selection and transition experiments, it is observed that the running gait has the lowest CoT. However, applying this gait is difficult due to the poor stability of existing control methods. To address this issue, we plan to design a new control method to improve the stability of the running gait on different terrains and implement our multi-gait strategy in real quadruped robots.

\addtolength{\textheight}{-1cm}   


\bibliographystyle{ieeetr}

\bibliography{zhang2024icra}

\begin{thebibliography}{10}

\bibitem{1CPG_gait_transition}
L.~Shang, Z.~Li, W.~Wang, and J.~Yi, ``Smooth gait transition based on cpg
  network for a quadruped robot,'' in {\em IEEE/ASME International Conference
  on Advanced Intelligent Mechatronics (AIM)}, pp.~358--363, 2019.

\bibitem{2bio_gait_transition}
I.~M. Koo, T.~D. Trong, Y.~H. Lee, and et~al., ``Biologically inspired gait
  transition control for a quadruped walking robot,'' {\em Autonomous Robots},
  vol.~39, pp.~169--182, 2015.

\bibitem{3Terrain_adaptive}
P.~Saraf, A.~Sarkar, and A.~Javed, ``Terrain adaptive gait transitioning for a
  quadruped robot using model predictive control,'' in {\em International
  Conference on Automation and Computing (ICAC)}, pp.~1--6, 2021.

\bibitem{4Autonomous_gait_transition}
T.~Fukui, H.~Fujisawa, K.~Otaka, and Y.~Fukuoka, ``Autonomous gait transition
  and galloping over unperceived obstacles of a quadruped robot with cpg
  modulated by vestibular feedback,'' {\em Robotics and Autonomous Systems},
  vol.~111, pp.~1--19, 2019.

\bibitem{5Automatic_gait_selection}
J.~Wang, I.~Chatzinikolaidis, C.~Mastalli, and et~al., ``Automatic gait pattern
  selection for legged robots,'' in {\em IEEE/RSJ International Conference on
  Intelligent Robots and Systems (IROS)}, pp.~3990--3997, 2020.

\bibitem{6Selecting_gaits}
W.~Xi, Y.~Yesilevskiy, and C.~D. Remy, ``Selecting gaits for economical
  locomotion of legged robots,'' {\em The International Journal of Robotics
  Research}, vol.~35, no.~9, pp.~1140--1154, 2016.

\bibitem{7smit2017energetic}
N.~Smit-Anseeuw, R.~Gleason, R.~Vasudevan, and C.~D. Remy, ``The energetic
  benefit of robotic gait selection—a case study on the robot ramone,'' {\em
  IEEE Robotics and Automation Letters}, vol.~2, no.~2, pp.~1124--1131, 2017.

\bibitem{lee2015gait}
Y.~H. Lee, D.~T. Tran, J.-h. Hyun, and et~al., ``A gait transition algorithm
  based on hybrid walking gait for a quadruped walking robot,'' {\em
  Intelligent Service Robotics}, vol.~8, pp.~185--200, 2015.

\bibitem{santos2011gait}
C.~P. Santos and Matos, ``Gait transition and modulation in a quadruped robot:
  A brainstem-like modulation approach,'' {\em Robotics and Autonomous
  Systems}, vol.~59, no.~9, pp.~620--634, 2011.

\bibitem{owaki2017quadruped}
D.~Owaki and A.~Ishiguro, ``A quadruped robot exhibiting spontaneous gait
  transitions from walking to trotting to galloping,'' {\em Scientific
  reports}, vol.~7, no.~1, p.~277, 2017.

\bibitem{quad_sdk}
J.~Norby, Y.~Yang, A.~Tajbakhsh, and et~al., ``Quad-sdk: Full stack software
  framework for agile quadrupedal locomotion,'' in {\em ICRA Workshop on Legged
  Robots}, 2022.

\bibitem{bledt2018cheetah}
G.~Bledt, M.~J. Powell, B.~Katz, and et~al., ``Mit cheetah 3: Design and
  control of a robust, dynamic quadruped robot,'' in {\em IEEE/RSJ
  International Conference on Intelligent Robots and Systems (IROS)},
  pp.~2245--2252, 2018.

\bibitem{ref-robust}
P.~Fankhauser, M.~Bjelonic, C.~D. Bellicoso, and et~al., ``Robust rough-terrain
  locomotion with a quadrupedal robot,'' in {\em IEEE International Conference
  on Robotics and Automation (ICRA)}, pp.~5761--5768, 2018.

\bibitem{ref-1986energy}
P.~Di~Prampero, ``The energy cost of human locomotion on land and in water,''
  {\em International journal of sports medicine}, vol.~7, no.~02, pp.~55--72,
  1986.

\bibitem{ref-hoyt1981gait}
D.~F. Hoyt and C.~R. Taylor, ``Gait and the energetics of locomotion in
  horses,'' {\em Nature}, vol.~292, no.~5820, pp.~239--240, 1981.

\bibitem{ref-shao2021learning}
Y.~Shao, Y.~Jin, X.~Liu, and et~al., ``Learning free gait transition for
  quadruped robots via phase-guided controller,'' {\em IEEE Robotics and
  Automation Letters}, vol.~7, no.~2, pp.~1230--1237, 2021.

\bibitem{ref-chen2022structural}
J.-P. Chen, H.-J. San, X.~Wu, and et~al., ``Structural design and gait research
  of a new bionic quadruped robot,'' {\em Proceedings of the Institution of
  Mechanical Engineers, Part B: Journal of Engineering Manufacture}, vol.~236,
  no.~14, pp.~1912--1922, 2022.

\bibitem{ref-sheng2022bio}
J.~Sheng, Y.~Chen, X.~Fang, and et~al., ``Bio-inspired rhythmic locomotion for
  quadruped robots,'' {\em IEEE Robotics and Automation Letters}, vol.~7,
  no.~3, pp.~6782--6789, 2022.

\bibitem{ref-kim2021gait}
J.~Kim, D.~X. Ba, H.~Yeom, and J.~Bae, ``Gait optimization of a quadruped robot
  using evolutionary computation,'' {\em Journal of Bionic Engineering},
  vol.~18, pp.~306--318, 2021.

\bibitem{ref-ctdi2018dynamic}
J.~Di~Carlo, P.~M. Wensing, B.~Katz, and et~al., ``Dynamic locomotion in the
  mit cheetah 3 through convex model-predictive control,'' in {\em IEEE/RSJ
  international conference on intelligent robots and systems (IROS)}, pp.~1--9,
  2018.

\bibitem{ref-fast}
J.~Norby and A.~M. Johnson, ``Fast global motion planning for dynamic legged
  robots,'' in {\em IEEE/RSJ International Conference on Intelligent Robots and
  Systems (IROS)}, pp.~3829--3836, 2020.

\bibitem{ref-ETH}
T.~Miki, J.~Lee, J.~Hwangbo, and et~al., ``Learning robust perceptive
  locomotion for quadrupedal robots in the wild,'' {\em Science Robotics},
  vol.~7, no.~62, p.~eabk2822, 2022.

\bibitem{ref-iit}
G.~Urbain, V.~Barasuol, C.~Semini, J.~Dambre, {\em et~al.}, ``Stance control
  inspired by cerebellum stabilizes reflex-based locomotion on hyq robot,'' in
  {\em IEEE International Conference on Robotics and Automation (ICRA)},
  pp.~6127--6133, 2020.

\bibitem{ref-design}
S.~Seok, A.~Wang, M.~Y. Chuah, and et~al., ``Design principles for highly
  efficient quadrupeds and implementation on the mit cheetah robot,'' in {\em
  IEEE International Conference on Robotics and Automation (ICRA)},
  pp.~3307--3312, 2013.

\bibitem{ref-nature}
Y.~Jin, X.~Liu, Y.~Shao, and et~al., ``High-speed quadrupedal locomotion by
  imitation-relaxation reinforcement learning,'' {\em Nature Machine
  Intelligence}, vol.~4, no.~12, pp.~1198--1208, 2022.

\bibitem{ref-biomechanical}
T.~M. Griffin, R.~Kram, S.~J. Wickler, and D.~F. Hoyt, ``Biomechanical and
  energetic determinants of the walk--trot transition in horses,'' {\em Journal
  of Experimental Biology}, vol.~207, no.~24, pp.~4215--4223, 2004.

\bibitem{intro1}
B.~R. Umberger and P.~E. Martin, ``Mechanical power and efficiency of level
  walking with different stride rates,'' {\em Journal of Experimental Biology},
  vol.~210, no.~18, pp.~3255--3265, 2007.

\bibitem{intro2}
A.~Muraro, C.~Chevallereau, and Y.~Aoustin, ``Optimal trajectories for a
  quadruped robot with trot, amble and curvet gaits for two energetic
  criteria,'' {\em Multibody System Dynamics}, vol.~9, no.~1, pp.~39--62, 2003.

\bibitem{intro3}
M.~Srinivasan and A.~Ruina, ``Computer optimization of a minimal biped model
  discovers walking and running,'' {\em Nature}, vol.~439, no.~7072,
  pp.~72--75, 2006.

\bibitem{ref-learning1}
Y.~Yang, T.~Zhang, E.~Coumans, J.~Tan, and B.~Boots, ``Fast and efficient
  locomotion via learned gait transitions,'' in {\em Conference on Robot
  Learning}, pp.~773--783, PMLR, 2022.

\bibitem{ref-learning2}
J.~Humphreys, J.~Li, Y.~Wan, H.~Gao, and C.~Zhou, ``Bio-inspired gait
  transitions for quadruped locomotion,'' {\em IEEE Robotics and Automation
  Letters}, vol.~8, no.~10, pp.~6131--6138, 2023.

\bibitem{APS}
Q.~Li, L.~Qian, S.~Wang, P.~Sun, and X.~Luo, ``Towards generation and
  transition of diverse gaits for quadrupedal robots based on trajectory
  optimization and whole-body impedance control,'' {\em IEEE Robotics and
  Automation Letters}, vol.~8, no.~4, pp.~2389--2396, 2023.

\end{thebibliography}

\end{document}